\ifdefined\XeTeXversion\else\pdfoutput=1\fi
\documentclass[11pt]{article}

\newif\ifreviewmode
 \reviewmodefalse

\ifreviewmode
  \usepackage[review]{acl}
\else
  \usepackage{acl}
\fi
\usepackage{times}
\usepackage{latexsym}
\usepackage{amssymb}
\usepackage[T1]{fontenc}
\usepackage[utf8]{inputenc}
\ifdefined\XeTeXversion
  \usepackage{newunicodechar}
  \newunicodechar{×}{\ensuremath{\times}}
  \newunicodechar{−}{\ensuremath{-}}
  \newunicodechar{≪}{\ensuremath{\ll}}
\else
  \DeclareUnicodeCharacter{00D7}{\ensuremath{\times}}
  \DeclareUnicodeCharacter{2212}{\ensuremath{-}}
  \DeclareUnicodeCharacter{226A}{\ensuremath{\ll}}
\fi
\usepackage{microtype}
\usepackage{inconsolata}
\usepackage{graphicx}
\usepackage{subcaption}
\usepackage{booktabs}
\usepackage{tabularx}
\usepackage{multirow}
\usepackage{array}
\usepackage{longtable}
\usepackage{enumitem}
\usepackage{placeins}
\usepackage{float}
\usepackage{amsmath}
\usepackage{xcolor}
\usepackage{colortbl}
\usepackage{pifont}
\usepackage{url}
\usepackage{verbatim}
\usepackage{mathtools}
\newcommand{\RC}{\textsc{Reasoning Core}}
\newcommand{\RG}{\textsc{Reasoning Gym}}
\newcommand{\PW}{\textsc{Procedural Warmup}}
\newcommand{\SL}{\textsc{SynLogic}}
\ifreviewmode
  \newcommand{\GitHubURL}{http://redacted}
  \newcommand{\HFURL}{http://redacted}
\else
  \newcommand{\GitHubURL}{https://github.com/sileod/reasoning-core}
  \newcommand{\HFURL}{https://hf.co/collections/reasoning-core/datasets}
\fi
\newcommand{\ReleaseLinks}{\footnote{Code:
  \href{\GitHubURL}{\includegraphics[height=1.25ex]{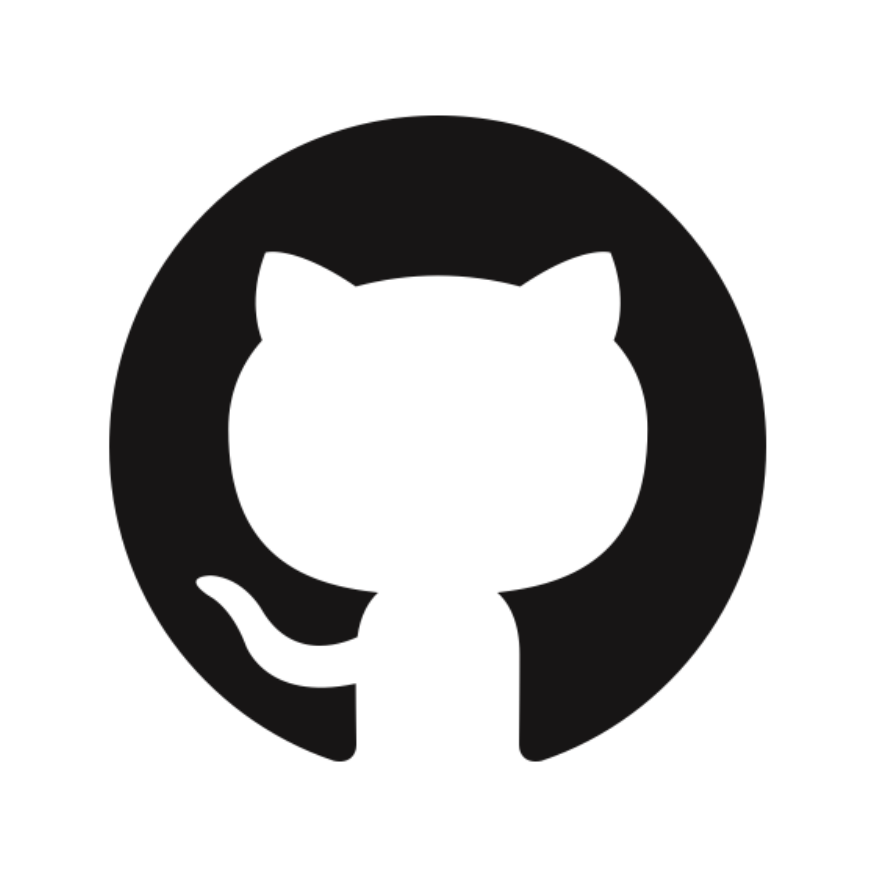}\,GitHub};
  data: \href{\HFURL}{\includegraphics[height=1.25ex]{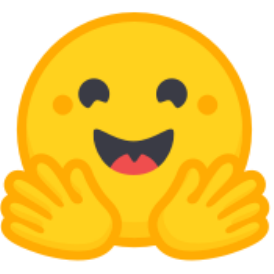}\,Hugging Face}.}}

\newcolumntype{Y}{>{\raggedright\arraybackslash}X}
\newcolumntype{C}[1]{>{\centering\arraybackslash}p{#1}}

\newcommand{\code}[1]{\texttt{\detokenize{#1}}}
\makeatletter
\newcommand{\inputwidetable}[1]{%
  \begingroup
  \expandafter\let\expandafter\table\csname table*\endcsname
  \expandafter\let\expandafter\endtable\csname endtable*\endcsname
  \input{#1}%
  \endgroup
}
\makeatother

\input{generated/claims}
\newcommand{\zsEvaluatedTasks}{50}
\newcommand{\zsHighBelowSeventyFive}{18}
\newcommand{\zsHighBelowNinety}{32}
\newcommand{\zsHighNearSaturated}{18}
\newcommand{\zsInstantMeanWhenHighSaturated}{0.61}
\newcommand{\zsProMeanWhenHighSaturated}{0.62}

\newcommand{\rlBbhTestCoreFinal}{38.6}
\newcommand{\rlBbhTestGymFinal}{23.1}
\newcommand{\rlMmluCoreFinal}{62.5}
\newcommand{\rlMmluGymFinal}{43.2}

\title{\RC{}: Designing Broad Procedural Data \\
for Completion-Supervised Reasoning Training}

\author{ Damien Sileo \and Valentin Lacombe \and Dimitri Kachler \\
Univ. Lille, Inria, CNRS, Centrale Lille, UMR 9189 - CRIStAL, F-59000 Lille, France \\
\texttt{damien.sileo@inria.fr}}

\begin{document}
\maketitle

\begin{abstract}
Procedural generators produce useful verifiable reasoning problems at scale, but
have received less attention as data for completion-supervised fine-tuning.
We introduce \RC{}, a collection of 50 generators spanning mathematics,
logic, planning, state tracking, formal languages, structured data, games,
causality, and code, with semantic scorers, difficulty controls, and task evaluators. Under a matched completion-supervised protocol, we compare \RC{} with \PW{}, \RG{}, and \SL{} across four base-model settings and multiple training durations. In the primary 3B comparison, \RC{} achieves the highest mean scores on DROP, LogiQA, and ARC-Challenge, exceeding both the baseline without procedural data and all three alternative procedural collections.

Task-level analyses show that semantic validity alone does not ensure training utility, highlighting compact targets and calibrated difficulty as important design factors. We ran audits combining model-assisted review, human adjudication, and regression testing. Applied throughout \RC{} development and to the other collections, they reveal subtle mismatches among generation, rendering, targets, and scoring, a reminder that procedural generation alone does not guarantee correctness.

The library, generated datasets, and audit material are publicly
available\ReleaseLinks{}.%
\end{abstract}

\section{Introduction}

The reasoning capabilities of language models are shaped by the problems they
encounter during training.  Procedural generators provide a powerful control over
that training distribution: they can produce fresh examples on demand, compute
or verify their targets, and vary structure and difficulty directly.  This
approach has been explored at different stages of training.
\citet{jiang2026proceduralpretrainingwarminglanguage} show that a narrow
warmup on abstract procedural sequences can improve subsequent language-model
training, while \RG{} and \SL{} provide broad collections of verifiable
reasoning environments, primarily for reinforcement learning with outcome
rewards
\citep{stojanovski2025reasoning,liu2025synlogic,r1}.

At the same time, a
rapidly growing literature has questioned how much reasoning reinforcement
learning creates beyond the capabilities and representations established
before RL, and has emphasized the importance of the supervised or earlier
training distribution \citep{Zhang2025OnTI,Akter2025FrontLoadingRT,yue2025doesreinforcementlearningreally,yuan2025fxgxfgxllms,matsutani2025rlsqueezessftexpands}.
Together, these results motivate studying the supervised-data layer directly:
can broad procedural collections strengthen the starting point on which later
training builds?

This leaves a basic data-design question: \emph{how should broad collections
of verifiable reasoning examples be designed and represented to become
effective supervised fine-tuning (SFT) data?}  
Verifiable generated problems are not automatically useful supervised data. Their value depends on whether the rendered instance determines the intended answer, how that answer is represented, which difficulty range provides a learnable signal, and whether the task remains useful in a heterogeneous mixture. Together, these choices determine how effectively a fixed token budget is converted into useful training signal.

We introduce \RC{}, a collection of 50 procedural generators designed for
broad completion-supervised training while retaining direct compatibility with
verifiable RL.  The tasks cover diverse domains (see Table \ref{tab:families}). %
They share a task interface, compact target conventions, semantic scoring,
difficulty controls, and a common diagnostic suite without forcing every task
into a single generation template.

We distinguish \emph{semantic validity}, whether a rendered instance has a
correct and identifiable target, from \emph{training utility}, whether exposure
improves downstream performance under a specified model, objective, dose, and
mixture. We study both through controlled comparisons of four procedural
collections across four base-model settings and several training durations,
complemented by free-generation, retention, and repository-audit diagnostics.

Our contributions are:
\begin{itemize}[leftmargin=*,nosep]
    \item \textbf{A broad, reproducible resource.} \RC{} provides 50
    generators with semantic scorers, difficulty controls, and one interface
    for SFT, evaluation, and RL. The release includes generated datasets,
    versioned manifests, caches, behavior hashes, task-level measurements, and
    audit material.
    \item \textbf{A controlled collection comparison.} We compare four
    procedural collections across four base-model settings and several training
    durations, with complete held-out evaluations.
    \item \textbf{A repository-scale semantic audit.} Model-assisted review,
    API-based sampling, manual adjudication, and regression testing expose
    subtle generator, target, and scorer failures in \RG{} and \SL{}.
\end{itemize}

\section{Related Work}

\paragraph{Verifiable environments and RL.}
Procedural content generation has long supported generalization studies in reinforcement learning \citep{cobbe2020leveraging,risi2020increasing}.
For language models, \RG{} offers a large collection of algorithmically scored environments \citep{stojanovski2025reasoning}, while \SL{} develops scalable logical games for RL with verifiable rewards \citep{liu2025synlogic}.
Related work builds verifiable tasks in games, theorem proving, logic, mathematics, and code \citep{balrog,lin2025zebralogicscalinglimitsllms,Xin2024DeepSeekProverATA,xu2025kodcode}.
Our study evaluates broad procedural collections under a completion-supervised objective rather than their native RL recipes, even though \RC{} scorers can nevertheless be used as outcome rewards.

\paragraph{Procedural supervision before RL.}
Formal languages, generated symbolic sequences, and abstract algorithmic tasks
have long been used to study how structured synthetic supervision shapes
learned representations and later language-model training
\citep{Wu2022InsightsIP,hu-etal-2025-circuits,allen2023physics}.
\PW{} consolidates previously proposed task families, including
formal-language problems such as Dyck tasks, into a broad public collection for
completion-supervised training
\citep{jiang2026proceduralpretrainingwarminglanguage}. This established
approach trains on compact, synthetically generated procedural completions
before broader language training. \RC{} extends the setting to a more
heterogeneous collection and evaluates generator-level design under matched
training interventions.

\paragraph{Curated instruction and reasoning supervision.}
Curated, non-procedural datasets remain a strong source of mathematical,
logical, conversational, and step-by-step reasoning supervision. DOLCI is a
public instruction-tuning suite containing human- and model-generated data for
step-by-step reasoning
(including GSM8K- and MATH-derived mathematical supervision), code,
precise instruction following, tool use, and high-quality conversation
\citep{olmo2025olmo}. Such resources assemble broad instruction-tuning
mixtures rather than generate structured task distributions from explicit
procedures.

\paragraph{Task utility estimation and data selection.}
Prior work estimates the influence of individual examples for targeted
instruction tuning, including using smaller models to select data for
larger ones \citep{pmlr-v235-xia24c}, or fits source-level mixture weights
from small-scale supervised fine-tuning experiments
\citep{pmlr-v267-li25bh}. We take a deliberately simpler approach at the
generator level: short, isolated SFT interventions on smaller models,
followed by direct measurement of task-rank agreement at larger scales.
We use this as a low-cost development diagnostic rather than as a
model-independent valuation method or an algorithm for optimizing the
training mixture.

\section{Collection design and validation}\label{sec:rc}

Building a broad procedural collection requires coordinated choices about which structures to sample, how to render them, how to represent and score targets, and how to integrate external solvers. We describe the final design here; Section \ref{sec:taskanalysis} reports the development evidence that informed several of these choices.

\begin{table*}[t]
\centering
\small
\begin{tabularx}{\textwidth}{@{}l C{0.04\textwidth} Y@{}}
\toprule
Family & \# & Representative tasks \\
\midrule
Mathematics & 6 & arithmetic expressions, word problems, equation systems, geometric relations, sequence induction, combinatorial formula selection \\
Formal proof and symbolic manipulation & 8 & Lean proof completion, Lean proof validation, Metamath entailment, Metamath premise selection, lambda reduction, term rewriting, unification, logical formalization \\
Logical and probabilistic inference & 8 & probabilistic evidence assignment, most-probable outcome prediction, entailment classification, multistep inference, defeasible inference, evidence retrieval, abduction, rule-system QA \\
Planning and games & 3 & plan generation, best-move selection, forced-win prediction \\
State and reference tracking & 4 & grid-state inference, reference tracking, belief tracking, coreference resolution \\
Graphs, constraints, and relations & 6 & shortest-path finding, multi-hop graph traversal, constraint solving, qualitative relation inference, causal reasoning, relational analogy \\
Formal languages and transduction & 6 & regex matching, equivalence and containment, parse derivation, syntax-error localization, constrained continuation, string transduction \\
Sets and structured data & 5 & missing-element retrieval, set-expression evaluation, typed table querying, table equivalence, statistical analysis \\
Code & 4 & program model checking, runnability classification, execution-result prediction, program synthesis \\
\bottomrule
\end{tabularx}
\caption{Coverage of the current \RC{} release. Appendix~\ref{sec:roster} describes all generators.}
\label{tab:families}
\end{table*}

\subsection{Designing procedural training data}

\paragraph{Broad, structured problem distributions.}
Because SFT uses a single teacher-forced  completion per example, it can cover many more distinct problems than rollout-based RL under a fixed compute budget. We seek broad problem distributions, varying semantic structure rather than merely surface form.
Where appropriate, task-specific recursive or grammar-based generators vary
expression trees, logical forms, program structure, and formal-language
systems.  For
planning and general game playing, we sample both problem instances and
underlying domains, including action schemas and transition rules for planning,
and rule systems and initial states for adversarial games. Generators therefore
need not repeatedly instantiate a fixed environment such as Blocksworld or
tic-tac-toe.

\paragraph{Difficulty control and forced sampling.}
Generators expose task-appropriate controls such as derivation depth, plan
length, variable count, branching factor, or the size of symbolic structures. A global difficulty knob maps a scalar level to these parameters through an overridable function, rather
than a fixed set of presets, enabling extrapolation.
High levels may become computationally expensive, but are calibrated so that levels up to 5 remain practical for every generator by convention. 

\paragraph{Rendering and prompt efficiency.}
We separate problem metadata from the prompt rendering making formatting a
flexible choice. The \RC{} prompts
used here provide zero-shot instructions without few-shot demonstrations, preserving more of the context and token budget for independently sampled problems. For external collections, we retain
original renderings, even though their tasks can have more complex answer formats and were not designed for SFT.%

\paragraph{SFT-aware answer design and semantic scoring.}
A semantically valid answer is not necessarily an effective supervised target.
We design answer serializations for the autoregressive objective, assigning a
deterministic canonical completion when suitable. This avoids arbitrary early
choices that induce incompatible but equally valid suffixes, particularly in
planning, proof construction, and program synthesis. For example, lexicographic ordering can canonicalize answers that are sets. We also favor compact
answers, such as a number, label, set, expression, proof-line index, plan, or
short program, over generated explanations. Under a fixed token budget, this exposes the model to more independently sampled problems and
reduces supervision devoted to incidental ordering or formatting.

Scoring remains broader than the supervised reference: sets are
order-invariant, lambda terms are compared up to variable renaming, plans are
executed, proof candidates are compiled, and generated programs are tested.
Training thus uses a canonical target while evaluation accepts
any semantically valid answer.

\paragraph{External tools and reproducibility.}
External solvers widen the range of checkable semantics but introduce
timeouts, version drift, malformed intermediate representations, unsafe
execution, and host-specific incompatibilities. We provide a common integration
layer and Docker images for the required solvers and runtimes. We pin
dependencies, isolate executable tasks, cache accepted instances, and separate
verifier failures from incorrect model answers. Generators and scorers are
tested independently so that verifier failures cannot yield credit.

\subsection{Task coverage and interface}

\RC{} contains 50 generators organized into nine broad groups
(Table~\ref{tab:families}).
The groups are descriptive rather than claims about independent cognitive skills.
Each task maps a seed and difficulty configuration to a prompt, a structured
reference answer, metadata, and a semantic scoring function.
These scorers support generator development, dataset validation,
free-generation evaluation, and online rewards.
The interface also supports \RG{} and \SL{}, enabling matched training and
evaluation across collections.
Most generators can produce effectively unbounded variations; experiment
manifests record exact samples and behavior hashes for reproducibility.

\subsection{Generator diagnostics}

Repository tests establish expected behavior on known cases, but they do not
characterize the generated training distribution.  We therefore use a
complementary set of diagnostics (Appendix Table~\ref{tab:diagnostics}) to
distinguish implementation defects, unlearnable difficulty ranges, shortcut
solutions, formatting costs, mixture redundancy, and model-specific effects.
These measurements guide generator changes and task analysis; they do not
define a universal automatic selection rule.

\subsection{Repository-scale semantic validation}\label{sec:semantic-audit}

Broad procedural repositories require semantic review beyond ordinary unit
tests. Throughout \RC{} development, repeated model-assisted reviews using
GPT-5.5 High and later GPT-5.6 High examined source code, generated prompts,
reference targets, and scorers. Separate API-based passes sampled rendered
examples using Claude Haiku 4.5 without reasoning at the lowest
task-difficulty settings (0 and 1), deliberately chosen to make errors easy to
identify and isolate. Cases receiving a reward other than 1 were flagged for
adjudication, initially using Claude Opus 4.6 and later Claude Opus 4.7,
followed by human review. Confirmed failures were manually reproduced, reduced
to minimal examples, and covered by regression tests. Model judgments were
never treated as ground truth.

Initial reviews of our tasks found subtle failures, including discrepancies between
conventional mathematical notation and Python arithmetic semantics, or ambiguous edge cases in logic tasks.
We iteratively corrected the tasks until no further  material defects were found in the final reviews.
To test whether the procedure could
expose consequential semantic failures rather than merely confirm our own
implementation, we applied the same audit to the other external collections used
in our experiments.

Under conservative criteria, we confirmed material default-path defects in 13
of 105 \RG{} tasks and nine native \SL{} generators. Examples include
a \RG{} scorer that awards full credit to any nonempty answer and a \SL{}
generator whose displayed constraints can contradict its stored solution.
Appendices ~\ref{sec:synlogic-code-audit} and ~\ref{sec:rg-code-audit} document
the audit scope and confirmed cases.

These findings are not aggregate judgments of either library. The comparison
is necessarily asymmetric: \RC{} was iteratively corrected during development,
whereas the external audits were retrospective and neither independent nor
blinded. We use the external audits to demonstrate the kinds of failures that
the validation procedure can expose, not to establish a quality ranking. The
results challenge the assumption that procedural data is correct by
construction: generation, rendering, targets, and scoring must remain aligned
across the repository.

\section{Experimental setup}
\label{sec:setup}

\subsection{Collections and training protocol}

We compare \RC{} with \PW{}, \RG{}, and \SL{} after converting all four
collections to prompt--answer examples trained with the same
completion-supervised objective. We reimplement \PW{} excluding the cellular automata task, which is a pure continuation task. %

We use four pretrained base models with publicly documented training corpora:
SmolLM2-135M,
SmolLM2-360M, %
OLMo-1B, and SmolLM3-3B-Base.
The main-data stream equally blends FineWeb-Edu \cite{penedo2024fineweb} text with DOLCI instruction and
conversation data. DOLCI includes step-by-step reasoning, tool-use, and
conversational supervision \citep{olmo2025olmo}; the main-only condition is a
curated SFT baseline rather than a raw-text baseline. The procedural collection
replaces a fraction of prompt-plus-answer tokens while the total number of
updates remains fixed.
We allocate 20\% of prompt-plus-answer tokens to auxiliary data.\footnote{We validated this ratio on \RG{} with a sweep over 10\%, 20\%, and 40\% using BBH validation set NLL at 2400 steps.}
We set a maximum sequence length of 1,024, and use paired seeds with a shared
data order within each seed. 
We evaluate
multiple durations from 300 to 2,400 updates.

All auxiliary conditions and the main-only baseline share the same seed and
optimization configuration. Reported effects are paired differences, so the
study isolates the marginal effect of auxiliary data. Learning rates were chosen in preliminary
\RG{} runs and then fixed across collections; no setting is tuned per collection.
Appendix Table~\ref{tab:sft-hparams} reports the complete configuration.

Tasks are sampled uniformly across supported levels and examples are deduplicated. Over-length rows are discarded rather than truncated because truncating from
the end would remove supervised answers.

\subsection{Held-out downstream evaluation}

Our primary measure is the relative reduction in held-out answer negative
log-likelihood from adding auxiliary collection \(A\):
\[
\Delta_{\mathrm{NLL}}(A;E)
\coloneqq
100\left(
1-
\frac{\operatorname{NLL}_E(M_{\mathrm{main}+A})}
     {\operatorname{NLL}_E(M_{\mathrm{main}})}
\right).
\]
Positive values indicate that adding \(A\) lowers held-out answer NLL on
evaluation distribution \(E\) relative to the matched main-only continuation.

\begin{figure*}[!t]
\centering
\begin{subfigure}{0.96\textwidth}
  \centering
\IfFileExists{figures/duration_ladder.pdf}{%
  \includegraphics[width=\linewidth]{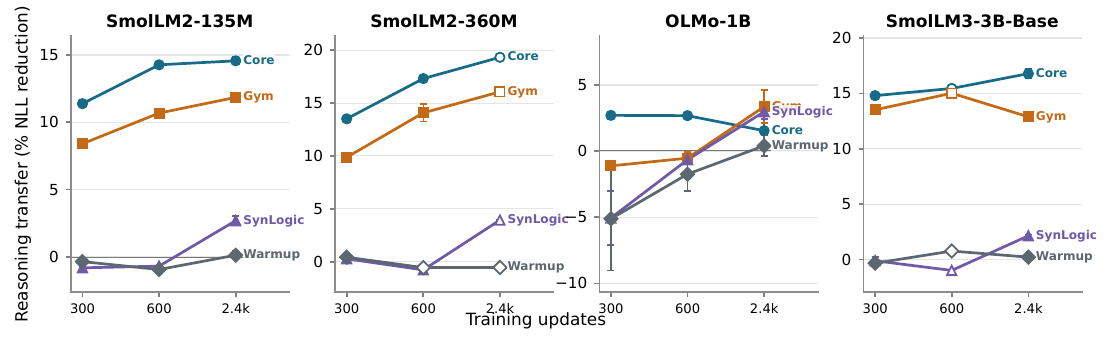}%
}{%
  \fbox{\parbox[c][3.2cm][c]{0.84\textwidth}{\centering Reasoning-transfer trajectories.}}%
}
\caption{Reasoning NLL compound over DROP, LogiQA, ARC-Challenge, and BBH-test}
\label{fig:duration_ladder}
\end{subfigure}

\begin{subfigure}{0.96\textwidth}
  \centering
\IfFileExists{figures/retention_ladder.pdf}{%
  \includegraphics[width=\linewidth]{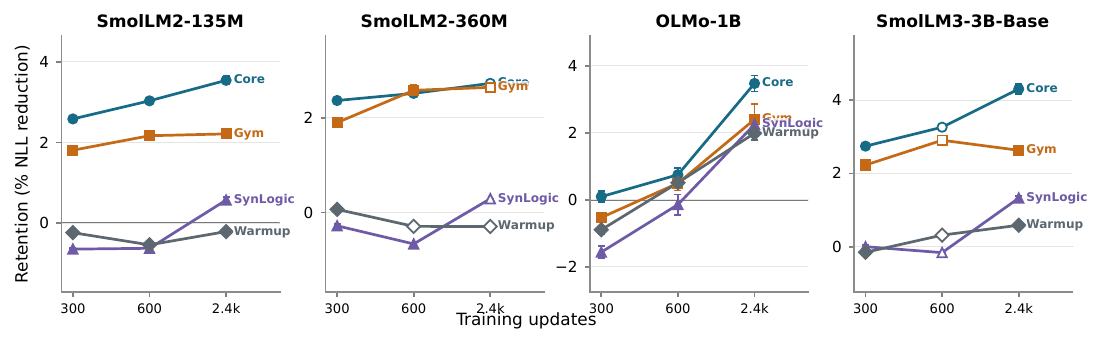}%
}{%
  \fbox{\parbox[c][3.2cm][c]{0.84\textwidth}{\centering Retention trajectories.}}%
}
\caption{Retention NLL compound over DOLCI and MMLU-other.}
\label{fig:retention_ladder}
\end{subfigure}
\caption{Transfer trajectories for the four procedural collections against
matched main-only continuations.  Values are relative answer-NLL reductions;
colored markers show means, and same-color error bars show mean $\pm$ standard
error over 3 seeds.
Panel-specific y-axis ranges emphasize within-model trends
and should not be compared across models.}
\label{fig:transfer_ladders}
\end{figure*}

The primary behavioral comparison uses two compounds. \emph{Reasoning} averages
over DROP, LogiQA, ARC-Challenge, and BBH-test. \emph{Retention} averages over
MMLU-other and DOLCI, and is deliberately weighted away from formal domains:
MMLU-other is a macro-average over 47 MMLU subjects after excluding mathematics,
formal logic, and computer science, while DOLCI measures NLL on the curated
instruction distribution the model is already training on. Retention therefore
asks whether procedural supervision is acquired at the expense of broader
capability. Figure~\ref{fig:transfer_ladders} reports the same two compounds as
relative answer-NLL reductions across training durations.

Accuracy is strongly quantized in most reported settings: exact match is near
floor at smaller scales, while multiple-choice evaluations remain near chance
and generally move by only 1--2 percentage points. We therefore use NLL wherever all held-out scores were logged, and report
accuracy and F1 at 3B (Table~\ref{tab:main_transfer_table}), where they become discriminative. NLL is paired and sensitive to partial probability
shifts at small scale. We additionally run task-native free-generation scoring
and zero-shot evaluations.

Answer NLL could in principle reward answer-format adaptation rather than
improved prediction. The Retention compound provides a control because DOLCI
targets are long-form, so a systematic shift toward short-answer serialization
would appear as a DOLCI NLL increase. Multiple-choice evaluations are scored
over full option text rather than compact labels, reducing any compact-format
advantage.

\begin{table*}[t]
\centering\footnotesize\renewcommand{\arraystretch}{0.92}\setlength{\tabcolsep}{2.5pt}
\begin{tabular}{lcccccc}
\toprule
 & \multicolumn{4}{c}{Reasoning} & \multicolumn{2}{c}{Retention} \\
\cmidrule(lr){2-5}\cmidrule(lr){6-7}
Auxiliary collection & DROP F1 $\uparrow$ & LogiQA Acc. $\uparrow$ & ARC-C Acc. $\uparrow$ & BBH-test $\uparrow$ & MMLU-other Acc. $\uparrow$ & DOLCI $\Delta$NLL $\downarrow$ \\
\midrule
Main-only & \cellcolor[RGB]{255,255,255}33.1{\scriptsize\color{black!55}\,$\pm$\,1.0} & \cellcolor[RGB]{255,255,255}46.8{\scriptsize\color{black!55}\,$\pm$\,0.6} & \cellcolor[RGB]{255,255,255}50.7{\scriptsize\color{black!55}\,$\pm$\,0.3} & \cellcolor[RGB]{255,255,255}43.6{\scriptsize\color{black!55}\,$\pm$\,1.1} & \cellcolor[RGB]{255,255,255}42.9{\scriptsize\color{black!55}\,$\pm$\,0.2} & \cellcolor[RGB]{255,255,255}0.000{\scriptsize\color{black!55}\,$\pm$\,0.000} \\
\RC{} & \cellcolor[RGB]{189,224,227}\textbf{41.7}{\scriptsize\color{black!55}\,$\pm$\,0.8} & \cellcolor[RGB]{189,224,227}\textbf{47.8}{\scriptsize\color{black!55}\,$\pm$\,0.7} & \cellcolor[RGB]{189,224,227}\textbf{51.3}{\scriptsize\color{black!55}\,$\pm$\,0.5} & \cellcolor[RGB]{189,224,227}\textbf{45.3}{\scriptsize\color{black!55}\,$\pm$\,1.1} & \cellcolor[RGB]{211,234,237}43.6{\scriptsize\color{black!55}\,$\pm$\,0.5} & \cellcolor[RGB]{209,233,235}$-$\,0.003{\scriptsize\color{black!55}\,$\pm$\,0.001} \\
\RG{} & \cellcolor[RGB]{203,230,233}39.2{\scriptsize\color{black!55}\,$\pm$\,2.4} & \cellcolor[RGB]{249,252,253}46.7{\scriptsize\color{black!55}\,$\pm$\,1.1} & \cellcolor[RGB]{207,232,235}51.1{\scriptsize\color{black!55}\,$\pm$\,0.2} & \cellcolor[RGB]{249,252,253}40.7{\scriptsize\color{black!55}\,$\pm$\,1.8} & \cellcolor[RGB]{189,224,227}\textbf{44.2}{\scriptsize\color{black!55}\,$\pm$\,0.3} & \cellcolor[RGB]{217,237,239}$-$\,0.002{\scriptsize\color{black!55}\,$\pm$\,0.001} \\
\SL{} & \cellcolor[RGB]{249,252,253}32.5{\scriptsize\color{black!55}\,$\pm$\,0.4} & \cellcolor[RGB]{229,243,244}47.1{\scriptsize\color{black!55}\,$\pm$\,0.5} & \cellcolor[RGB]{249,252,253}49.7{\scriptsize\color{black!55}\,$\pm$\,0.5} & \cellcolor[RGB]{249,252,253}41.6{\scriptsize\color{black!55}\,$\pm$\,0.9} & \cellcolor[RGB]{236,246,247}42.9{\scriptsize\color{black!55}\,$\pm$\,0.2} & \cellcolor[RGB]{189,224,227}\textbf{$-$\,0.006}{\scriptsize\color{black!55}\,$\pm$\,0.000} \\
Proc.\ Warmup & \cellcolor[RGB]{249,252,253}31.3{\scriptsize\color{black!55}\,$\pm$\,0.5} & \cellcolor[RGB]{205,231,234}47.7{\scriptsize\color{black!55}\,$\pm$\,0.4} & \cellcolor[RGB]{249,252,253}49.1{\scriptsize\color{black!55}\,$\pm$\,0.3} & \cellcolor[RGB]{249,252,253}42.3{\scriptsize\color{black!55}\,$\pm$\,0.9} & \cellcolor[RGB]{249,252,253}42.1{\scriptsize\color{black!55}\,$\pm$\,0.3} & \cellcolor[RGB]{194,226,230}$-$\,0.005{\scriptsize\color{black!55}\,$\pm$\,0.001} \\
\bottomrule
\end{tabular}

\caption{SmolLM3-3B-Base after 2,400 updates across 5 seeds. Scores are
percentage points, except DOLCI $\Delta$NLL. Darker teal indicates larger
improvements over main-only; gray values are sample standard deviations.}
\label{tab:main_transfer_table}
\end{table*}

\subsection{Development data}

We partition BBH \cite{suzgun-etal-2023-challenging} using the task taxonomy
provided by its authors: non-algorithmic NLP tasks form the development set,
while algorithmic tasks are reserved for held-out testing.  The development
subset is less directly aligned with the procedural generators studied here,
reducing the risk of selecting tasks that merely reproduce the evaluation
structure. No \RC{} generator instantiates, adapts, or reproduces a BBH-development task.
BBH and FineWeb NLL were the only validation signals used during development.  Task-development decisions were based on runs of 300 steps with SmolLM2-360M and OLMo-1B only.
\section{Results}
\label{sec:results}

\subsection{Collections comparison on downstream tasks}

At 3B and 2,400 updates, \RC{} improves all Reasoning metrics over the
paired main-only continuation. Across models and durations, each procedural
collection yields positive reasoning transfer in at least some settings, but
their relative ordering varies. 
Figure~\ref{fig:transfer_ladders} shows the Reasoning and Retention trajectories across training durations. 

Appendix Table \ref{tab:additional_nll} reports the individual NLL reductions. We additionally evaluate a 50-task Reasoning Gym subset selected by BBH-development NLL and sampled uniformly. The subset improves LogiQA but degrades most other metrics relative to the full collection, suggesting that BBH-development utility does not transfer uniformly across evaluations and that narrowing the task mixture can reduce aggregate transfer.

Table~\ref{tab:main_transfer_table} reports the comparison on individual datasets and confirms that NLL gains convert to accuracy improvements.

Appendix Table~\ref{tab:duration_rank_correlations}
reports collection-order agreement over all four collections.

\subsection{Analyzing tasks and task design}
\subsubsection{Task-level agreement within collections}

For task development, we compare isolated-task ranks across the three base
models with complete matched panels; \PW{} has only a collection-level
intervention.  This diagnostic is explicitly developmental: it retains BBH
development in its score and is not part of the held-out collection comparison.
Within the SmolLM2 family, Kendall's $\tau_b$
is +0.68 for \RC{}, +0.75 for \RG{}, and +0.88 for \SL{}, preserving 84\%,
88\%, and 94\% of pairwise task orderings, respectively
(Table~\ref{tab:task_rank_agreement}).  Thus, the 135M panel supplies a direct
development signal for task changes intended for the same model family.

\begin{table}[t]\centering\scriptsize\setlength{\tabcolsep}{3.5pt}
\caption{Within-collection task-rank agreement in the task-development analysis. Each cell reports Kendall's $\tau_b$ and the percentage of preserved pairwise orderings under the BBH-development diagnostic score for isolated 300-update interventions.}\label{tab:task_rank_agreement}
\begin{tabular}{lccc}
\toprule
Model pair & Core ($n$=50) & Gym ($n$=95) & SynLogic ($n$=18) \\
\midrule
135M--360M & +0.68 (84\%) & +0.75 (88\%) & +0.88 (94\%) \\
135M--OLMo & +0.34 (67\%) & +0.44 (72\%) & -0.24 (35\%) \\
360M--OLMo & +0.38 (69\%) & +0.39 (70\%) & -0.26 (34\%) \\
\bottomrule
\end{tabular}
\end{table}

\subsubsection{Learnability, format, and transfer}
\label{sec:taskanalysis}
We next analyze isolated-task runs observationally, correlating task properties with transfer; \RG{} provides the largest such panel and spans very different learning regimes. After 300 updates, SmolLM2-360M reaches at least .90 native reward on 6 of 95 tasks, including \code{graph_color} and \code{propositional_logic}, while 45 remain at or below .05; \code{count_primes} stays at zero.

Native reward is informative but incomplete. Tasks making little progress are generally weak on BBH, and reward gain is more informative than initial reward. Within Reasoning Gym, intermediate final rewards show the strongest held-out transfer, while fully saturated tasks do not transfer better. Saturation may reflect mastery of a narrow output space rather than broadly useful learning.

Grid and board tasks form a conspicuous negative cluster. They often pair a long state description with a short decision or a fully serialized solution. On FineWeb, their apparent penalty is mostly explained by prompt and answer length. Long-input decisions can still help BBH while hurting retention, whereas full-state reconstruction tends to hurt both. Some constraint grids, including sudoku and futoshiki, remain weak on BBH after controlling for length, suggesting a task-specific mismatch rather than a general failure of spatial reasoning.

Token matching separates this length effect from auxiliary dose. Across 18 paired OLMo-1B tasks, it improves BBH in every case, but leaves the relationship between example length and FineWeb retention unchanged. Long examples therefore remain costly even when the auxiliary token budget is controlled.
Full task-level estimates and controls are reported separately
(Appendix~\ref{app:task-diagnostics}).

\subsubsection{A negative result on step-by-step rationale targets}

We also tested step-by-step targets for tasks including parsing and graph pathfinding. For parsing, the targets reproduced the successive operations of the Earley chart parser. For graph pathfinding, they traced breadth-first search on unweighted graphs and Dijkstra’s algorithm on weighted graphs, including frontier expansion, distance updates, and predecessor choices. These rationales were semantically correct, but they consistently performed worse than compact-answer supervision on both BBH development and FineWeb NLL. The same ordering held with matched training budgets and without budget matching, and making the traces shorter and more token-efficient did not reverse it.  In these tasks, the more transferable skill may be learning the gestalt of the problem: seeing how a string fits together under a grammar, or how a viable route takes shape in a graph, rather than reproducing the solver’s bookkeeping one operation at a time. The result does not imply that rationale supervision is generally harmful. It does show, however, that a faithful transcript of a correct algorithm can be a worse training target than the answer it computes.
\subsection{Zero-shot solvability}

Training utility and initial solvability are different axes. A task already
solved by frontier models may remain useful for evaluation but provide little
headroom, while low zero-shot reward may indicate a useful teacher-forced
signal, malformed data, or excessive difficulty. We therefore run task-native
free-generation evaluation at multiple difficulty levels.
Appendix Figure~\ref{fig:zeroshot} evaluates the roughly 300B-parameter
DeepSeek V4 Flash in Instant and High reasoning modes and the roughly
1.6T-parameter DeepSeek V4 Pro in Instant mode. Flash
Instant$\rightarrow$High probes inference-time reasoning scaling with the model
fixed, while Flash$\rightarrow$Pro probes parameter scaling. These evaluations
were not used for generator or task selection. High accuracy under increased
reasoning effort does not make a task redundant for completion-supervised
training, since its examples may still improve direct prediction or the
representations supporting longer reasoning.

Flash High remains below 0.75 reward on
\zsHighBelowSeventyFive{}/50 evaluated tasks and below 0.90 on
\zsHighBelowNinety{}/\zsEvaluatedTasks{}. Among the
\zsHighNearSaturated{} tasks where Flash High reaches at least 0.90, Flash
Instant and Pro Instant achieve mean rewards of
\zsInstantMeanWhenHighSaturated{} and \zsProMeanWhenHighSaturated{},
respectively. Thus, neither the 300B Flash nor the 1.6T Pro configuration fully
solves the collection zero-shot. Increased inference-time reasoning solves more
tasks, while both Instant configurations retain substantial headroom. Several
tasks near-saturated by Flash High remain difficult for both Instant
configurations.  The within-model Instant-to-High contrast is more
consistently positive than the contrast between models in Instant mode.

\subsection{Verifier-backed RL with \RC{}}
\label{sec:rlcomparison}

\begin{figure}
\centering
\includegraphics[width=\columnwidth]{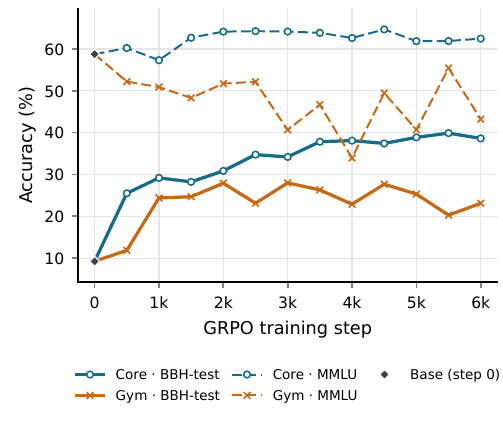}
\caption{\textbf{Reasoning Core vs Reasoning Gym under matched RL.}
BBH-test (solid) and MMLU (dashed) accuracy during GRPO training of
Qwen2.5-3B-Instruct. Colors distinguish the two procedural collections.
The two single-seed conditions share the training configuration and differ only
in the procedural collection and corresponding verifier.}
\label{fig:rlcomparison}
\end{figure}

We compare \RC{} and \RG{} in a matched, single-seed GRPO run on
Qwen2.5-3B-Instruct following \citet{stojanovski2025reasoning}. Both conditions
use the same optimizer, sampling configuration, effective batch size, and
200k-example budget; only the training collection and verifier differ.
Figure~\ref{fig:rlcomparison} reports BBH-test and MMLU accuracy every 500
steps through step 6,000. At the final checkpoint, \RC{} reaches
\rlBbhTestCoreFinal\% versus \rlBbhTestGymFinal\% on BBH-test and
\rlMmluCoreFinal\% versus \rlMmluGymFinal\% on MMLU. However, the \RG{} run
degrades MMLU below the base model, suggesting run-specific instability rather
than a reliable collection-level difference. We therefore treat the gap as
illustrative only. The narrower conclusion is that \RC{} can support
verifier-backed RL.
\section{Discussion}

\paragraph{Procedural collections are not tied to a single training objective.}
\RG{} and \SL{} were developed primarily as verifiable RL environments, yet
both provide useful completion-supervised training data under our protocol,
with \RG{} performing particularly strongly.
Conversely, \RC{} was designed around completion-supervised data utility but
also supports effective verifier-backed RL.
However, orienting design choices explicitly toward the completion objectives leads to better overall SFT results.
Completion-supervised training consumes generated problems directly, rather than
generating and scoring several on-policy rollouts for each prompt, and can
therefore expose the model to more independently sampled problems under a
fixed compute budget.
Our experiments evaluate completion supervision and RL separately; whether
the same procedural collection should be reused for sequential SFT and RL
remains open.

\paragraph{Task utility depends on the training context.}
A generator does not have a single model-independent value.
Its measured effect depends on the base model, target representation, training
dose, main distribution, and neighboring tasks.
Isolated-task interventions are therefore most useful for diagnosing and
revising generators under settings close to those in which they were measured.
Collection-level comparisons answer a different question: whether a complete
procedural mixture supplies useful marginal supervision after task-specific
effects and interactions are combined.

\section{Conclusion}
\RC{} provides 50 verifiable generators under a common interface for supervised
training, evaluation, and RL. In the primary 3B comparison, \RC{} improves all
reported Reasoning metrics over the shared main-only continuation; across models
and durations, every procedural collection produces positive reasoning transfer
in at least some settings, while relative ordering varies across metrics, model
families, and training durations. Target design matters as much as task
semantics: compact canonical answers outperformed faithful step-by-step traces
of the correct algorithm, and audits confirmed that procedural generation alone
does not guarantee correctness. We release roughly 10B generated tokens under
permissive licenses (Appendix~\ref{app:data-release}); the released diagnostics
keep these conclusions inspectable at the generator level and support future
SFT--RL comparisons.

\section*{Limitations}
Although collection-level transfer does not appear to deteriorate across the model sizes studied, our evidence stops at 3B parameters, so whether these trends persist at larger scales remains unknown; model-family coverage is also limited.
The evaluation emphasizes formal and closed-answer reasoning; it does not establish improvements on open-ended, multimodal, or real-world agentic tasks.
Although the shared DOLCI baseline already supplies curated reasoning
supervision, we do not compare against matched mixtures of its individual
human-curated components, such as GSM8K or MATH, nor do we evaluate formal-logic benchmarks beyond
the MMLU and BBH scores used here.
We evaluate completion-supervised training and verifier-backed RL as separate
interventions. We do not test sequential SFT followed by RL on the same
procedural collection, so the experiments do not determine how procedural SFT
changes subsequent RL or whether reusing the same collection across both stages
is beneficial. The matched RL comparison uses a single seed and should be
interpreted as a demonstration of viability rather than a precise estimate of
the collection-level margin.

\ifreviewmode\else
\section*{Acknowledgements}
This work was supported by the French National Research Agency (ANR) through the ANR-24-CE23-4637 grant (Adada project).

Experiments presented in this paper were carried out using the Grid'5000 testbed, supported by a scientific interest group hosted by Inria and including CNRS, RENATER, and several universities as well as other organizations (see \url{https://www.grid5000.fr}).
\fi

\bibliography{custom}

\clearpage
\onecolumn
\raggedbottom
\appendix
\section{List of \RC{} tasks}
\label{sec:roster}

\newcommand{\taskdesc}[2]{%
  \noindent\textbf{\texttt{#1}}\quad #2%
  \par\vspace{2pt}%
}
\taskdesc{arithmetics}{
Expression trees sampled from a grammar combine arithmetic, rounding,
number-theoretic, divisor-related, prime-related, counting, and bit-level
operations. Instances are evaluated using either exact rational arithmetic
or Python semantics, with a cue in the prompt disambiguating the intended
semantics when there is a difference.
}

\taskdesc{math\_word\_problem}{
Constructively sampled process chains and relational quantity graphs
combine forward arithmetic, inverse operations, fractions, and aggregate
constraints. Symbolic solving retains instances with a unique admissible
interpretation.
}

\taskdesc{equation\_system}{
Latent integer solutions are used to construct mixed linear systems.
Generation covers uniquely solvable, inconsistent, and underdetermined
regimes, with symbolic algebra providing the oracle.
}

\taskdesc{lean\_missing\_line}{
Generated Lean theorem and proof schemas are converted into incomplete
proofs. Candidate lines are compiled individually, and instances are
retained when exactly one candidate completes the proof.
}

\taskdesc{lean\_candidate\_compilation}{
Valid Lean proofs are paired with targeted proof mutations. Each
candidate is checked through the Lean REPL, making compilation success
the direct semantic oracle.
}

\taskdesc{planar\_geometry\_relations}{
Exact rational planar scenes combine primitive and derived points,
intersections, projections, reflections, rotations, and affine
transformations. Geometric predicates are verified symbolically.
}

\taskdesc{metamath\_entailment}{
Proof fragments are mined from \texttt{set.mm} and extended with fresh
substitution trees. An independent stack verifier checks entailment,
while ablations and semantic foils produce controlled alternatives.
}

\taskdesc{metamath\_core\_select}{
Candidate theorem sets are constructed around a verified Metamath proof.
Premise ablation identifies a minimal sufficient core rather than merely
a set containing the correct theorem.
}

\taskdesc{combinatorics\_formula\_selection}{
Latent counting programs represent selections, distributions,
arrangements, unions, lattice paths, role assignments, and constrained
strings. Each program is compiled into several formulas, including
distractors encoding common counting misconceptions.
}

\taskdesc{lambda\_reduction}{
Lambda terms are sampled directly or generated by verified
anti-reduction from normal forms. Capture-avoiding reduction is checked
against the target normal form up to alpha-equivalence.
}

\taskdesc{rewrite\_system}{
Rewrite problems are generated from normal forms by inserting redexes.
Evaluation follows deterministic rule and position priority, allowing
multi-step interaction among independently generated rewrite rules.
}

\taskdesc{unification\_entailment}{
Terms are generated from shared latent structures and then partially
masked, renamed, or augmented with distractors. A unification engine
computes the most general unifier and verifies each proposed consequence.
}

\taskdesc{most\_probable\_evidence}{
Boolean formulas and hidden probabilistic factors define compact
probabilistic programs. ProbLog computes the maximum-probability evidence
assignment under the generated constraints.
}

\taskdesc{most\_probable\_outcome}{
Probabilistic experiment programs compose random choices, conditional
branches, and derived outcomes. Exact program evaluation determines the
most probable observable result.
}

\taskdesc{logic\_nli}{
Formal semantic structures and their natural-language realizations are
generated jointly. An automated theorem prover distinguishes entailment,
contradiction, and neutrality while also checking premise consistency.
}

\taskdesc{logic\_formalization}{
English statements and TPTP formulas are generated from common semantic
structures. Meaning-preserving and meaning-breaking mutations are
classified through automated logical comparison.
}

\taskdesc{multistep\_nli}{
Horn-style theories are constructed with controlled proof depth,
branching, and support size. Positive, negative, and unknown conclusions
are determined from the generated deductive closure.
}

\taskdesc{defeasible\_nli}{
Rules combine explicit negation, defaults, exceptions, and stratified
negation-as-failure. Instances require reasoning about both ordinary
deduction and the defeat of otherwise applicable defaults.
}

\taskdesc{multistep\_evidence\_retrieval}{
A conclusion is embedded in a larger generated theory containing
irrelevant premises. Systematic ablation identifies the premises that
are individually necessary for the target derivation.
}

\taskdesc{multistep\_abduction}{
Generated theories contain missing facts required for a target
conclusion. Candidate additions are enumerated and compared to find
minimum-cardinality abductive explanations.
}

\taskdesc{logic\_qa}{
Questions are posed over the closure of a generated rule system.
Objectives include retrieving derived entities, testing membership, and
computing simple aggregates over logically inferred facts.
}

\taskdesc{planning}{
Fresh typed planning domains contain generated objects, fluents,
operators, preconditions, effects, initial states, and goals. Exact
search finds a shortest valid plan, with a stable action-order rule used
when several shortest plans exist.
}

\taskdesc{set\_missing\_element}{
An element is removed from an ordered generated universe and the
remaining elements are shuffled. Evaluation parses answers
order-insensitively and verifies the omitted value exactly.
}

\taskdesc{set\_expression}{
Random abstract syntax trees compose set and list operations, predicates,
counts, and duplicate-bearing intermediate lists. Direct execution of
the generated expression supplies the answer.
}

\taskdesc{sequential\_induction}{
Sequences are produced by a bounded polynomial recurrence language over
the index, earlier terms, constants, and arithmetic operators.
Enumeration selects a uniquely simplest recurrence consistent with the
visible prefix.
}

\taskdesc{qualitative\_reasoning}{
Instances draw from Allen interval relations, cardinal directions,
RCC8, and ordinal rankings. Composition tables, path consistency, exact
constraint solving, and clue minimization support multi-hop qualitative
inference.
}

\taskdesc{grid\_navigation}{
Random collision-free grid worlds evolve through moves, swaps, and
jumps. A time-indexed constraint model determines coordinates, Manhattan
distances, or spatial relations only when the queried result is uniquely
entailed.
}

\taskdesc{reference\_tracking}{
Entities undergo moves, swaps, bulk transfers, and descriptions using
pronouns or indirect references. Optional Winograd-style cases use
verified relational chains before the final location is queried.
}

\taskdesc{belief\_tracking}{
Explicit event semantics model nested beliefs under public, private,
one-way, and missed observations. Reports, deception, attribution,
failed communication, and counterfactual event pairs create controlled
epistemic divergences.
}

\taskdesc{coreference}{
Variable-width lineups evolve through forward and backward
permutations, branching possibilities, later evidence, and distractors.
Equality, difference, permutation, and all-different constraints are
solved jointly.
}

\taskdesc{constraint\_satisfaction}{
A shared finite-domain generator produces assignment, graph, scheduling,
grid, set, and numeric CSPs. Queries cover unique values, possibility,
relations, counterfactuals, consistency, complete solution sets, and
lexicographic optima.
}

\taskdesc{graph\_pathfinding}{
Directed or undirected, weighted or unweighted graphs are sampled from
several random topology families and rendered in varied textual forms.
Shortest paths are checked semantically, including disconnected cases
and deterministic resolution of equal-length alternatives.
}

\taskdesc{graph\_successors}{
Random permutation digraphs give every node exactly one successor.
Queries require following the transition relation for an arbitrary
number of hops rather than inspecting a single edge.
}

\taskdesc{regex\_following}{
Regexes are sampled from a compositional grammar. Search identifies the
shortest nonempty visible ASCII string that fully matches the expression,
using lexical order only to resolve equal-length matches.
}

\taskdesc{regex\_reasoning}{
Generated regular expressions are compiled into finite-state machines.
Automata operations test equivalence and containment or construct a
shortest witness in the symmetric difference.
}

\taskdesc{analogical\_case\_matching}{
Random relational structures are linked by injective object and
predicate mappings, with selected predicates reversed. Matching requires
transporting the full relation structure while rejecting hard negative
and distractor cases.
}

\taskdesc{parsing\_derivation}{
Fresh productive context-free grammars generate uniquely parsed strings.
An Earley parser recovers the structure, and the target is expressed as
the corresponding sequence of shuffled rule labels in leftmost order.
}

\taskdesc{syntax\_error\_detection}{
Grammar-generated strings are kept valid, truncated, or corrupted by
insertion, deletion, or substitution. An exact parser-derived oracle
locates the first invalid position or the expected continuation.
}

\taskdesc{constrained\_continuation}{
A generated sentence contains a missing multi-token span. Parser states and
grammar constraints enumerate valid completions, and instances are retained
only when the intended continuation is unique.
}

\taskdesc{table\_qa}{
Typed tables contain numbers, dates, booleans, strings, and nulls and are
serialized in several formats. Generated SQL combines filters,
expressions, aggregation, grouping, ordering, and limits, with DuckDB
execution serving as the oracle.
}

\taskdesc{table\_equivalence}{
Tables are compared after normalizing typed column identifiers and
interpreting rows as a multiset. Row order, column order, syntax, and
formatting may vary, while duplicate-row multiplicity remains
semantically significant.
}

\taskdesc{table\_statistics}{
Generated tables support Pearson correlation, eta squared, normalized
mutual information, partial correlation, group robustness,
heterogeneity, and distribution-shift questions. Margin constraints
remove numerically ambiguous cases.
}

\taskdesc{string\_transduction}{
String transformations are generated either as small programs or as
explicit edit scripts. Operations include reversal, sorting,
deduplication, rotation, Caesar shifts, replacement, filtering,
insertion, deletion, and substitution.
}

\taskdesc{game\_best\_move}{
Random finite directed-acyclic games are compiled into Game Description
Language and executed with a Clingo-based interpreter. Complete minimax
evaluation retains positions with a unique, nondegenerate best move.
}

\taskdesc{game\_forced\_win}{
The same generated game framework is used to determine whether the
current player can force an outcome above a specified utility threshold.
Degenerate and severely imbalanced instances are filtered out.
}

\taskdesc{qualitative\_causal\_reasoning}{
Causal kernels represent direct and mediated effects, blocked paths,
competing paths, confounders, colliders, explaining away, multiple
treks, and observe-versus-intervene contrasts. Additional random DAG
edges are accepted only when they preserve the intended answer.
}

\taskdesc{code\_analysis}{
Grammar-generated finite-state Python-like programs contain categorical,
integer, and Boolean variables, branches, guards, pattern matches, tuple
updates, and nondeterministic choices. Their transition systems are
model-checked against generated CTL formulas.
}

\taskdesc{code\_runnability}{
Typed multi-function Python programs are paired with targeted AST
mutations that induce specific runtime failures. Isolated execution with
resource and step limits determines whether the program succeeds and,
when it fails, the exact exception class.
}

\taskdesc{code\_execution}{
Fresh procedurally generated typed Python functions are executed in a restricted
multiprocessing sandbox. The target is the exact returned
representation, with filters removing trivial constant, identity, or
excessively short executions.
}

\taskdesc{program\_synthesis}{
Programs are enumerated in a typed string-fragment DSL containing
concatenation, substrings, replacement, conditionals, arithmetic, search,
and predicates. Counterexample-guided example selection eliminates
cheaper hypotheses, while held-out cases validate the unique
minimum-cost program.
}

\section{Zero-shot Solvability}
\label{sec:appendix-zeroshot}

\begingroup
\setlength{\intextsep}{0pt}
\begin{figure}[H]
  \vspace{-0.75\baselineskip}
  \centering
  \includegraphics[width=\linewidth]{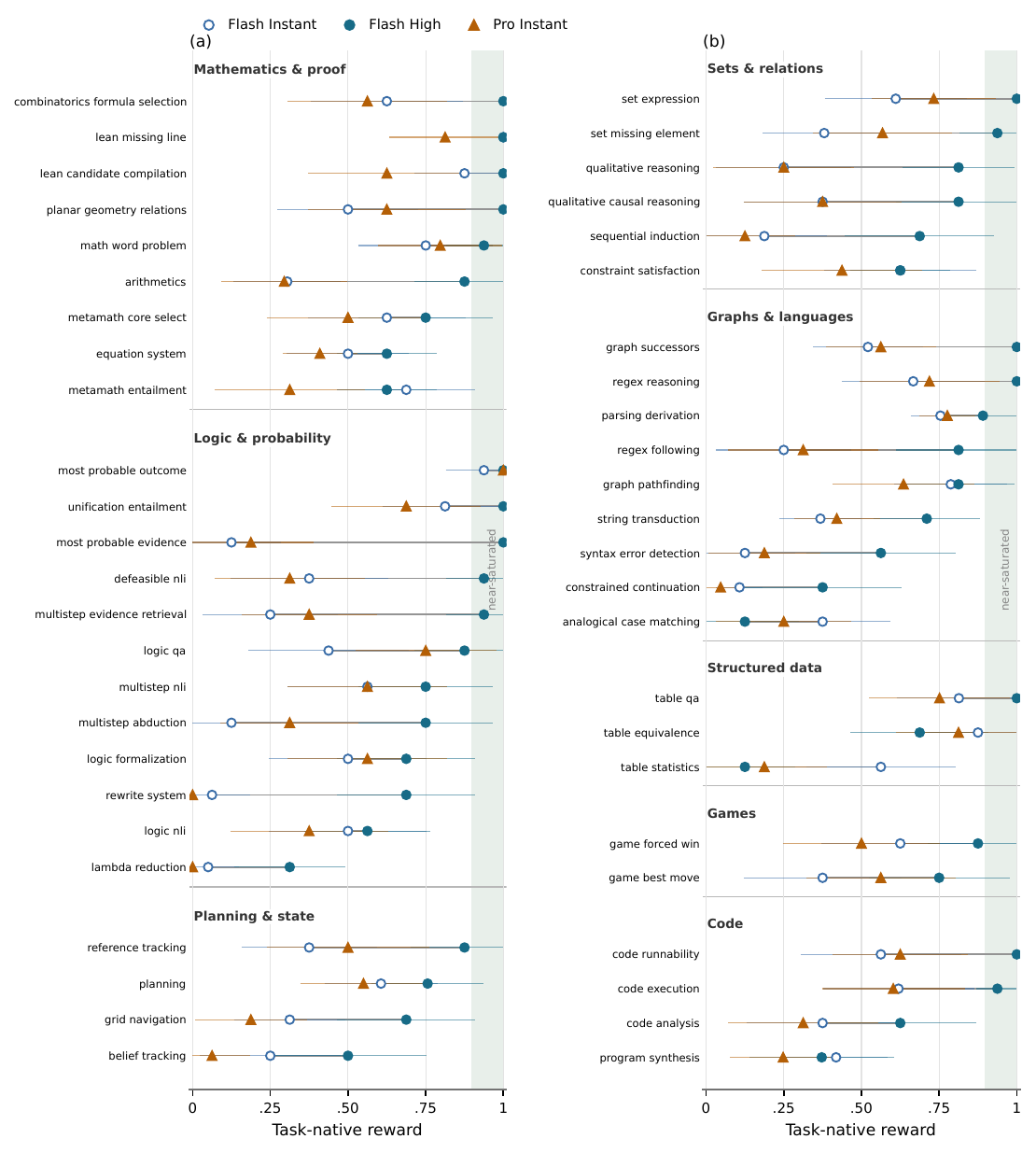}
  \caption{Zero-shot task-native reward across the evaluated \RC{} tasks.
  Each row reports DeepSeek V4 Flash Instant, Flash High, and Pro Instant; tasks
  are grouped by family and ordered by Flash High reward. Lines connect the two
  Flash configurations and therefore show the effect of increased inference-time
  reasoning within the same model. Thin bars show stratified 95\% intervals
  from within-difficulty sampling variation. The shaded region denotes
  near-saturation (reward $\geq 0.90$). }
  \label{fig:zeroshot}
\end{figure}
\endgroup

\section{A Diagnostic Checklist for Procedural Supervision}
\begin{table}[H]
\centering
\small
\begin{tabularx}{\linewidth}{@{}p{0.25\linewidth} Y Y@{}}
\toprule
Check & Evidence & Action \\
\midrule
Do the prompt, reference, and scorer agree?
& consistent disagreement between strong models and the reference, followed by human-confirmed contradictions or scoring errors
& fix the generator or scorer and add a regression test \\

Is the output unique or canonical?
& multiple correct answers, unresolved ties, equivalent answers with different serializations
& impose a deterministic convention, such as lexicographic tie-breaking, or compare answers semantically \\

Is the answer distribution varied and free of shortcuts?
& dominant labels, low target entropy, or prompt-only baselines based on class prior, words, length, or formatting perform substantially above chance
& rebalance or diversify outputs, remove predictive cues, or discard the variant \\

Does the scorer reject incorrect outputs?
& wrong labels, malformed answers, nonsense, or small adversarial mutations receive nonzero reward
& test known-valid, known-invalid, malformed, and adversarial candidates; make verifier failures fail closed \\

Is the difficulty range learnable?
& task-native reward collapses, training accuracy remains near zero, or downstream utility becomes harmful at higher levels
& restrict the default range to the learnable region \\

Is the target token-efficient?
& long targets reduce examples per token budget and correlate with weaker marginal utility
& shorten or canonicalize the target; measure prompt and answer lengths separately \\

Does the task add value in the mixture?
& isolated gains disappear when related tasks are combined or oversampled
& reduce redundant sampling and evaluate marginal utility in the intended mixture \\

Is utility stable across training regimes?
& task effects change rank or sign across model families, sizes, or durations
& report regime-specific results rather than a universal task ranking \\
\bottomrule
\end{tabularx}
\caption{Checklist for diagnosing implementation defects and limitations of task design, representation, and mixture composition.}
\label{tab:diagnostics}
\end{table}

\paragraph{Automated coverage.}
Every generator exposes a \texttt{validate()} method that performs common
interface-level smoke tests. It checks that prompts are non-empty and sampled
examples are non-degenerate, that stored reference answers receive full credit,
that unrelated sampled answers are not universally accepted, and that malformed
or empty candidates do not crash the scorer. It additionally checks scoring
after metadata serialization, difficulty-level mutability, and that generation
does not reset global randomness. These checks cover only basic
generator--scorer plumbing; they do not establish semantic correctness, target
uniqueness, shortcut resistance, learnability, or mixture utility.

\FloatBarrier
\clearpage

\section{Additional results \label{app:additional}}

\subsection{Per-benchmark NLL transfer}

\begin{table}[H]
\centering\footnotesize\renewcommand{\arraystretch}{0.92}\setlength{\tabcolsep}{2pt}
\resizebox{\linewidth}{!}{\begin{tabular}{lccccccc}
\toprule
 & \multicolumn{5}{c}{Reasoning} & \multicolumn{2}{c}{Retention} \\
\cmidrule(lr){2-6}\cmidrule(lr){7-8}
Auxiliary collection & DROP & LogiQA & ARC-E & ARC-C & BBH-test & MMLU-other & DOLCI \\
\midrule
\RC{} & \cellcolor[RGB]{189,224,227}\textbf{23.39}{\scriptsize\color{black!55}\,$\pm$\,1.62} & \cellcolor[RGB]{204,231,233}4.40{\scriptsize\color{black!55}\,$\pm$\,0.99} & \cellcolor[RGB]{189,224,227}\textbf{13.21}{\scriptsize\color{black!55}\,$\pm$\,1.17} & \cellcolor[RGB]{189,224,227}\textbf{9.21}{\scriptsize\color{black!55}\,$\pm$\,0.84} & \cellcolor[RGB]{198,228,231}29.17{\scriptsize\color{black!55}\,$\pm$\,2.42} & \cellcolor[RGB]{189,224,227}\textbf{8.06}{\scriptsize\color{black!55}\,$\pm$\,0.62} & \cellcolor[RGB]{209,233,235}0.41{\scriptsize\color{black!55}\,$\pm$\,0.08} \\
\RG{} & \cellcolor[RGB]{195,227,230}20.33{\scriptsize\color{black!55}\,$\pm$\,2.99} & \cellcolor[RGB]{217,237,239}2.57{\scriptsize\color{black!55}\,$\pm$\,0.89} & \cellcolor[RGB]{215,236,238}5.95{\scriptsize\color{black!55}\,$\pm$\,1.58} & \cellcolor[RGB]{218,237,239}3.54{\scriptsize\color{black!55}\,$\pm$\,1.12} & \cellcolor[RGB]{205,231,234}23.89{\scriptsize\color{black!55}\,$\pm$\,2.50} & \cellcolor[RGB]{207,232,235}4.99{\scriptsize\color{black!55}\,$\pm$\,0.29} & \cellcolor[RGB]{217,237,239}0.28{\scriptsize\color{black!55}\,$\pm$\,0.07} \\
\RG{} (filtered) & \cellcolor[RGB]{198,228,231}19.11{\scriptsize\color{black!55}\,$\pm$\,2.21} & \cellcolor[RGB]{202,230,233}4.64{\scriptsize\color{black!55}\,$\pm$\,0.84} & \cellcolor[RGB]{223,240,242}3.71{\scriptsize\color{black!55}\,$\pm$\,1.56} & \cellcolor[RGB]{226,241,243}2.08{\scriptsize\color{black!55}\,$\pm$\,1.00} & \cellcolor[RGB]{206,232,234}23.40{\scriptsize\color{black!55}\,$\pm$\,1.97} & \cellcolor[RGB]{209,233,236}4.70{\scriptsize\color{black!55}\,$\pm$\,0.66} & \cellcolor[RGB]{216,236,239}0.30{\scriptsize\color{black!55}\,$\pm$\,0.06} \\
Core/Gym equal mix & \cellcolor[RGB]{199,228,231}18.48{\scriptsize\color{black!55}\,$\pm$\,2.05} & \cellcolor[RGB]{189,224,227}\textbf{6.37}{\scriptsize\color{black!55}\,$\pm$\,0.64} & \cellcolor[RGB]{206,232,234}8.50{\scriptsize\color{black!55}\,$\pm$\,1.40} & \cellcolor[RGB]{208,232,235}5.59{\scriptsize\color{black!55}\,$\pm$\,0.86} & \cellcolor[RGB]{189,224,227}\textbf{36.25}{\scriptsize\color{black!55}\,$\pm$\,1.86} & \cellcolor[RGB]{203,230,233}5.73{\scriptsize\color{black!55}\,$\pm$\,0.34} & \cellcolor[RGB]{197,228,231}0.57{\scriptsize\color{black!55}\,$\pm$\,0.07} \\
\SL{} & \cellcolor[RGB]{232,244,245}2.06{\scriptsize\color{black!55}\,$\pm$\,0.43} & \cellcolor[RGB]{221,239,241}2.09{\scriptsize\color{black!55}\,$\pm$\,0.36} & \cellcolor[RGB]{230,243,244}1.81{\scriptsize\color{black!55}\,$\pm$\,0.53} & \cellcolor[RGB]{229,243,244}1.32{\scriptsize\color{black!55}\,$\pm$\,0.35} & \cellcolor[RGB]{233,244,246}2.84{\scriptsize\color{black!55}\,$\pm$\,0.52} & \cellcolor[RGB]{225,241,242}1.96{\scriptsize\color{black!55}\,$\pm$\,0.25} & \cellcolor[RGB]{189,224,227}\textbf{0.69}{\scriptsize\color{black!55}\,$\pm$\,0.05} \\
Proc.\ Warmup & \cellcolor[RGB]{249,252,253}-1.41{\scriptsize\color{black!55}\,$\pm$\,0.83} & \cellcolor[RGB]{230,243,244}0.87{\scriptsize\color{black!55}\,$\pm$\,0.41} & \cellcolor[RGB]{249,252,253}-0.74{\scriptsize\color{black!55}\,$\pm$\,0.67} & \cellcolor[RGB]{249,252,253}-0.47{\scriptsize\color{black!55}\,$\pm$\,0.43} & \cellcolor[RGB]{234,245,246}2.01{\scriptsize\color{black!55}\,$\pm$\,0.77} & \cellcolor[RGB]{233,244,246}0.56{\scriptsize\color{black!55}\,$\pm$\,0.35} & \cellcolor[RGB]{194,226,230}0.61{\scriptsize\color{black!55}\,$\pm$\,0.06} \\
\bottomrule
\end{tabular}
}
\caption{SmolLM3-3B-Base after 2,400 updates with 20\% auxiliary tokens.
Each cell reports the paired mean relative NLL reduction from adding auxiliary
data versus the main-only continuation at the same seed and configuration,
with sample standard deviation in gray; higher is better. DOLCI measures
retention on the training distribution.
\RG{} "(filtered)" denotes the 50 best tasks from the original collections according to BBH validation scores after 300 steps with Olmo 1B.}
\label{tab:additional_nll}
\end{table}

\subsection{Collection ordering across durations}
\begin{table}[H]\centering\small
\begin{tabular}{lccccc}
\toprule
Earlier $\rightarrow$ later & 135M & 360M & OLMo-1B & SmolLM3-3B & Mean \\
\midrule
300 $\rightarrow$ 600 & 0.67 (5/6) & 1.00 (6/6) & 1.00 (6/6) & 0.67 (5/6) & 0.83 \\
300 $\rightarrow$ 1,200 & 0.67 (5/6) & 0.67 (5/6) & 0.00 (3/6) & 0.33 (4/6) & 0.42 \\
600 $\rightarrow$ 1,200 & 1.00 (6/6) & 0.67 (5/6) & 0.00 (3/6) & 0.67 (5/6) & 0.58 \\
600 $\rightarrow$ 2,400 & 1.00 (6/6) & 0.67 (5/6) & 0.33 (4/6) & 0.67 (5/6) & 0.67 \\
\bottomrule
\end{tabular}
\caption{Collection-order agreement across training durations. Each cell reports Kendall's $\tau_b$ and, in parentheses, the number of preserved pairwise orderings out of six. The mean averages $\tau_b$ over models with complete cells for all four procedural collections.}\label{tab:duration_rank_correlations}
\end{table}

\subsection{Task-native learnability}
\begin{table}[H]\centering\small
\caption{Task-native free-generation reward at the end of training (learnability diagnostic)}\label{tab:saturation_reward_table}
\begin{tabular}{lcccc}
\toprule
Auxiliary collection & 135M & 360M & OLMo-1B & SmolLM3-3B \\
\midrule
\RC{} & 0.025 & 0.130 & 0.269 & 0.345 \\
\RG{} & 0.024 & 0.069 & 0.130 & 0.217 \\
\SL{} & 0.030 & 0.013 & 0.034 & 0.034 \\
\bottomrule
\end{tabular}
\end{table}

\subsection{Task-level learnability and format diagnostics}
\label{app:task-diagnostics}

\paragraph{Setup.}
We analyze the isolated 300-update \textsc{Reasoning Gym} interventions for SmolLM2-360M and OLMo-1B. Each intervention mixes one auxiliary task with the shared main stream and is compared with the main-only continuation using the same seed and optimization configuration. For auxiliary task \(t\) and evaluation distribution \(E\), we define its influence as
\[
I_E(t)
\coloneqq
\Delta_{\mathrm{NLL}}(\{t\};E).
\]

Task-native reward is measured before and after training, with reward gain defined as their difference. Prompt and answer lengths are averaged over deterministic samples stratified by task and difficulty level. We standardize task features and influence within each base model before pooling the panels. Correlations are Pearson correlations with 95\% intervals obtained by resampling tasks. Grid tasks are identified from the task definition; adjusted estimates control jointly for mean prompt and answer length. The held-out reasoning compound averages influence on DROP, LogiQA, ARC-Challenge, and BBH-test.

\begin{table}[t]
\centering
\footnotesize
\setlength{\tabcolsep}{3pt}
\begin{tabular}{@{}p{0.61\columnwidth}p{0.33\columnwidth}@{}}
\toprule
Diagnostic & Estimate [95\% interval] \\
\midrule
\multicolumn{2}{@{}l}{\textit{Learnability}} \\
SmolLM2-360M reward $\geq .90$ / $\leq .05$
    & $6/95$ / $45/95$ \\
BBH vs.\ initial / final / gained reward
    & $.13$ / $.23$ / $.20$ \\
Intermediate vs.\ saturated final reward, held-out reasoning
    & $+.48$ SD $[-.01,.94]$ \\
\addlinespace
\multicolumn{2}{@{}l}{\textit{Length and format}} \\
FineWeb vs.\ prompt / answer length
    & $-.68$ / $-.27$ \\
BBH vs.\ prompt / answer length
    & $-.09$ / $-.27$ \\
Grid effect on FineWeb, raw / adjusted
    & $-.26$ / $+.03$ \\
Grid effect on BBH, raw / adjusted
    & $-.25$ / $-.21$ \\
\addlinespace
\multicolumn{2}{@{}l}{\textit{Token-matching control}} \\
BBH change from token matching
    & $+1.77$ pp $[1.15,2.48]$; $18/18$ improve \\
FineWeb length $\rho$, example / token matched
    & $-.85$ / $-.87$ \\
\bottomrule
\end{tabular}
\caption{Task-level diagnostics underlying Section~\ref{sec:taskanalysis}.
Correlations pool within-model standardized panels. Raw and adjusted grid
effects denote correlations before and after controlling for prompt and answer
length.}
\label{tab:task-diagnostics}
\end{table}

\paragraph{Token-matching control.}
The paired OLMo-1B comparison uses the same 18 tasks, task pool, seed, main stream, and 300-update schedule. Example matching samples auxiliary examples with probability $.20$, corresponding to 13.1\% of training tokens because auxiliary examples are shorter than main examples. Token matching raises the example probability to $.293$ to obtain the intended 20\% auxiliary-token share. Differences are computed within task.

\FloatBarrier

\clearpage
\section{Hyperparameters, reproducibility and licensing}

\subsection{Fine-tuning hyperparameters}

\begin{table}[H]
\centering
\small
\begin{tabularx}{\linewidth}{@{}p{0.25\linewidth}Y@{}}
\toprule
Setting & Value \\
\midrule
Framework & TRL \texttt{SFTTrainer} 1.2.0; Transformers 4.57.6; PyTorch 2.7.1 (CUDA 12.6); Python 3.12 \\
Optimizer & AdamW (\texttt{adamw\_torch}), $\beta_1=0.9$, $\beta_2=0.999$, $\epsilon=10^{-8}$ \\
Learning rate & $10^{-4}$ (135M, 360M); $2\times10^{-5}$ (OLMo-1B, SmolLM3-3B) \\
Schedule / warmup & Linear decay to zero; no warmup \\
Weight decay / clipping & 0.01; max-norm 1.0 \\
Effective batch & 8 sequences in all runs; device batch $\times$ accumulation $8\times1$ ($\leq$360M) and $8/k \times k$ for $k\in\{2,4,8\}$ ($\geq$1B), with $k$ set per accelerator to fit memory and the product held fixed at 8 \\
Checkpointing / precision & On ($\geq$1B), off ($\leq$360M); bf16 mixed precision \\
Packing / sequence length & Multi-sequence packing; 1,024 tokens \\
Loss / mixture & Completion-only masking; 80\% main / 20\% auxiliary tokens interleave, first-exhausted, per-seed shuffle \\
\bottomrule
\end{tabularx}
\caption{SFT configuration.  Within each base-model setting, every collection and its paired main-only baseline use identical optimization settings.}
\label{tab:sft-hparams}
\end{table}

\subsection{Zero-shot hyperparameters}
Requests to DeepSeek V4 Flash and DeepSeek V4 Pro were routed through OpenRouter to the DeepInfra provider without overriding generation parameters. For both models, OpenRouter's defaults were used: \verb|temperature=1|, \verb|top_p=1|, \verb|top_k=0|, \verb|min_p=0|, \verb|frequency_penalty=0|, \verb|presence_penalty=0|, and \verb|repetition_penalty=1|. Outputs were capped at 8192 tokens.

\paragraph{Cached data and manifests.}
Every reported intervention uses cached task examples and a manifest recording generator version, behavior hash, model, seed, main distribution, auxiliary ratio, optimizer settings, and evaluation components.
A shared main-only baseline is reused only when all corresponding configuration fields match.

\paragraph{Over-length handling.}
Rows for which the prompt, answer, and end-of-sequence token together exceed the 1,024-token budget are discarded before packing.
They are never clipped, because clipping the end of a sequence would disproportionately remove answer tokens under completion-only supervision.
The evaluator records a sentinel for discarded examples so that per-example outputs remain aligned with benchmark and subject identifiers.

\newcommand{\PP}{\textsc{Procedural Pile}}
\subsection{Pre-generated data and code release}
\label{app:data-release}
We release \PP{}, approximately \(10\) billion prompt-and-answer tokens
generated from \textsc{Reasoning Core}, together with substantial generated
slices of \textsc{Reasoning Gym}, \textsc{SynLogic}, and
\textsc{Procedural Warmup} in a common format. The release provides broad
reusable coverage rather than a prescribed training budget. \PP{} is
distributed under CC BY 4.0; redistributed slices retain the licenses of their
source collections. The code is released under an MIT license.

\clearpage

\newcommand{\slcommit}{d8c527fd17edb739172619efb9b681805fc74b8d}
\newcommand{\slcommitlink}{%
\href{https://github.com/MiniMax-AI/SynLogic/commit/\slcommit}%
{\texttt{d8c527f}}%
}
\newcommand{\slcode}[2]{%
\href{https://github.com/MiniMax-AI/SynLogic/blob/\slcommit/#1}%
{\texttt{#2}}%
}

\section{SynLogic Semantic Audit}
\label{sec:synlogic-code-audit}

\begingroup
\scriptsize
\setlength{\tabcolsep}{4pt}
\renewcommand{\arraystretch}{1.08}

\begin{longtable}{
>{\raggedright\arraybackslash}p{3.35cm}
>{\raggedright\arraybackslash}p{11.15cm}}

\caption{
Major semantic defects found among the native \textsc{SynLogic} task
generators at commit \slcommitlink, ordered by severity. We include only
failures affecting default generation: incorrect or non-entailed labels,
incomplete targets for tasks requesting all valid answers, and material
mismatches between the rendered problem and the computation used to produce
the oracle. Valid but non-unique witnesses are listed separately and are not
counted as semantic defects. Generator names link to the audited source code.
}
\label{tab:synlogic-semantic-errors}\\

\toprule
Generator & Major semantic defect \\
\midrule
\endfirsthead

\multicolumn{2}{l}{%
\scriptsize\itshape
Table~\ref{tab:synlogic-semantic-errors} continued
}\\
\toprule
Generator & Major semantic defect \\
\midrule
\endhead

\midrule
\multicolumn{2}{r}{\scriptsize Continued on next page}\\
\endfoot

\bottomrule
\endlastfoot

\multicolumn{2}{l}{\textbf{Directly invalid or non-entailed labels}}\\
\addlinespace[2pt]

\slcode{games/tasks/star_placement_puzzle/scripts/star_placement_puzzle.py}
{star\_placement\_puzzle}
& The default \(4\times4\), one-star configuration is hard-coded as
\(\{(0,0),(1,3),(2,1),(3,2)\}\). The stars at \((2,1)\) and \((3,2)\)
are diagonally adjacent, directly violating the stated rule that stars may
not touch, including diagonally. This hard-coded branch returns without
calling the generator's own placement validator, so default examples can
store an invalid solution. \\

\slcode{games/tasks/futoshiki/scripts/futoshiki_generator.py}
{futoshiki}
& The generator first samples a Latin square and then assigns each displayed
inequality sign independently and uniformly from \(\{<,>\}\), without
checking the values in the corresponding cells. Each sign therefore has
probability \(1/2\) of contradicting the stored grid. With the default four
inequalities, only \(1/16=6.25\%\) of generated labels are expected to
satisfy every displayed constraint. \\

\slcode{games/tasks/object_properties/scripts/object_properties.py}
{object\_properties}
& In an exchange transformation, the rendered description specifies only
the size and material of the replacement objects. The implementation also
samples hidden names, origins, smells, and colours for those objects and
later answers property queries from this unmentioned internal state.
Generated answers about these hidden attributes are therefore not entailed
by the question and should instead be unknown. \\

\slcode{games/tasks/calcudoko/scripts/calcudoko.py}
{calcudoko}
& Division-cage targets are computed with integer floor division rather than
requiring exact divisibility. Values such as \(3\) and \(2\) consequently
produce a target of \(1\div\), although neither exact quotient equals \(1\).
In addition, when a single cell remains after region construction, it is
appended to the preceding cage without recomputing that cage's operator or
target. The stored grid can therefore violate the displayed cage
constraints. \\

\slcode{games/tasks/time_sequence/scripts/time_sequence.py}
{time\_sequence}
& The special rule stating that a participant is in a time zone one hour
ahead is not implemented as a shift of that participant's schedule.
Instead, the code merely adds a busy interval covering the final hour of
the workday. Whenever this rule is sampled, the availability calculation
and resulting label describe a different scheduling problem from the one
shown in the prompt. \\

\slcode{games/tasks/operation/scripts/operation.py}
{operation}
& The expression evaluator reduces operators of equal precedence
left-associatively, including exponentiation, whereas the rendered
\texttt{**} operator conventionally associates to the right. It can thus
assign the wrong value to expressions containing exponentiation chains.
Conditional custom operations can also take the default branch incorrectly:
an intermediate integral float such as \texttt{2.0} fails the
\texttt{int}-string check, so its stated integer condition is not evaluated. \\

\slcode{games/tasks/boolean_expressions/scripts/boolean_expressions.py}
{boolean\_expressions}
& The supposedly true factual leaves include the claim that the Great Wall
of China is visible from space. Under the ordinary unaided-eye
interpretation this is false, or at minimum interpretation-dependent.
Boolean expressions whose value depends on this leaf are consequently
labelled using an unstated and generally incorrect factual premise. \\

\addlinespace[4pt]
\multicolumn{2}{l}{\textbf{Incomplete requested targets}}\\
\addlinespace[2pt]

\slcode{games/tasks/minesweeper/scripts/minesweeper.py}
{minesweeper}
& The prompt asks for all unrevealed cells that must be mines, but the oracle
implements only the direct saturation rule in which a revealed number's
remaining mine count equals its number of unknown neighbours. It omits
general constraint and subset reasoning, so additional logically forced
mines may be absent from the target. The purported uniqueness check does not
establish completeness: after marking the directly discovered mines, it
returns true whenever at least one such mine was found. \\

\addlinespace[4pt]
\multicolumn{2}{l}{\textbf{Material prompt--trace mismatch}}\\
\addlinespace[2pt]

\slcode{games/tasks/word_sorting_mistake/scripts/word_sorting_mistake.py}
{word\_sorting\_mistake}
& In character-reading error branches, the generator mutates the
\texttt{words} list that is subsequently used to render the source word
list. The reasoning trace nevertheless uses a reverse map to display the
pre-mutation spelling while computing letters and sorting states from the
mutated token. The rendered input and the annotated reasoning can therefore
refer to different word sets, so the labelled first-error step is not
defined relative to a single fixed problem instance. \\

\end{longtable}
\endgroup

\begingroup\footnotesize

\noindent\textbf{Valid but non-canonical targets.}
The following generators appear to construct valid witnesses but do not
establish that the rendered instance has a unique solution:
\texttt{arrow\_maze},
\texttt{campsite},
\texttt{kukurasu},
\texttt{math\_path},
\texttt{norinori},
\texttt{number\_wall},
\texttt{numbrix},
\texttt{skyscraper\_puzzle},
\texttt{survo}, and
\texttt{wordscapes}.
We do not count this as a major semantic defect because these tasks generally
admit verifier-based evaluation of alternative valid witnesses. It remains a
data-design concern when training against one stored completion with exact
answer supervision.

\noindent\textbf{Non-default semantic defect.}
The basic default mode of
\slcode{games/tasks/web_of_lies/scripts/web_of_lies.py}{web\_of\_lies}
appears correct. In its extended mode, however, the false rendering of
``at least one person tells the truth'' is ``at least one person lies.''
These propositions are not logical negations: both hold when one person is
truthful and the other lies. Extended-mode labels can therefore be computed
from semantics different from the displayed statement.

\endgroup

\newcommand{\rgcommit}{49b07130b3fcd12f2d064bba7c43869543a0e7e7}
\newcommand{\rgcommitlink}{%
\href{https://github.com/open-thought/reasoning-gym/commit/\rgcommit}%
{\texttt{49b0713}}%
}
\newcommand{\rgcode}[2]{%
\href{https://github.com/open-thought/reasoning-gym/blob/\rgcommit/reasoning_gym/#1.py}%
{\texttt{#2}}%
}
\clearpage

\section{Reasoning Gym Semantic Audit}
\label{sec:rg-code-audit}

\begingroup
\scriptsize
\setlength{\tabcolsep}{4pt}
\renewcommand{\arraystretch}{1.08}

\begin{longtable}{
>{\raggedright\arraybackslash}p{3.35cm}
>{\raggedright\arraybackslash}p{11.15cm}}

\caption{
Major semantic and evaluation defects found among the 105 registered
\textsc{Reasoning Gym} generators at commit \rgcommitlink, ordered by
severity. We include failures affecting default generation or scoring:
scorers that accept incorrect outputs, incorrect or non-entailed labels,
and material mismatches between the stated task and the implemented
semantics. Underdetermined targets scored against a single sampled answer
are listed separately after the table and, as for \textsc{SynLogic}, are not
counted as semantic defects. Generator names link to the audited source code.
}
\label{tab:reasoning-gym-semantic-errors}\\

\toprule
Generator & Major semantic or evaluation defect \\
\midrule
\endfirsthead

\multicolumn{2}{l}{%
\scriptsize\itshape
Table~\ref{tab:reasoning-gym-semantic-errors} continued
}\\
\toprule
Generator & Major semantic or evaluation defect \\
\midrule
\endhead

\midrule
\multicolumn{2}{r}{\scriptsize Continued on next page}\\
\endfoot

\bottomrule
\endlastfoot

\multicolumn{2}{l}{\textbf{Scoring and verification failures}}\\
\addlinespace[2pt]

\rgcode{algorithmic/game_of_life_halting}{game\_of\_life\_halting}
& The scorer converts both the candidate and reference strings with
\texttt{bool}. Since the reference is always the nonempty string
\texttt{"True"} or \texttt{"False"}, every nonempty candidate, including
the wrong label or arbitrary text, receives full credit. \\

\rgcode{algebra/intermediate_integration}{intermediate\_integration}
& One default branch constructs undefined symbolic functions named
\texttt{sin} and \texttt{cos}. SymPy may consequently retain an unevaluated
integral as the oracle, while the derivative-based verifier accepts that
oracle tautologically because differentiating the unevaluated integral
reconstructs the original integrand. \\

\rgcode{games/boxnet}{boxnet}
& The prompt requires conflict-free, efficient multi-agent plans, but the
evaluator applies actions sequentially, silently skips invalid actions, and
does not evaluate simultaneous-action conflicts or plan efficiency. Reward
depends only on the fraction of boxes that disappear after the accepted
actions. \\

\rgcode{games/tsumego}{tsumego}
& The generator plants a candidate capturing move and then adds a random decoy
stone. It subsequently checks only that the stored move remains legal, not
that it still captures any stone. The scorer nevertheless exact-matches
against that move, without checking whether another move captures more
stones or whether the stored move is still a capture at all. \\

\addlinespace[4pt]
\multicolumn{2}{l}{\textbf{Incorrect or non-entailed labels}}\\
\addlinespace[2pt]

\rgcode{games/mahjong}{mahjong\_puzzle}
& A desired outcome is sampled before the added card is selected.
\texttt{Chi} candidates are not rechecked for the higher-priority
\texttt{Peng} condition, while fallback cases are labelled \texttt{Pass}
without checking either rule. Default generation can therefore assign
directly false outcome labels. \\

\rgcode{logic/syllogisms}{syllogism}
& The validity checker tests distribution of the conclusion terms but omits
the requirement that the middle term be distributed in at least one
premise. Invalid categorical syllogisms exhibiting the undistributed-middle
fallacy are consequently labelled valid. \\

\rgcode{graphs/family_relationships}{family\_relationships}
& When the second person is a parent of the first person's spouse, the
implementation returns the relationship of the second person to the first.
The prompt asks for the reverse direction, causing daughter- or son-in-law
relations to be labelled mother- or father-in-law. \\

\rgcode{arithmetic/calendar_arithmetic}{calendar\_arithmetic}
& Business-day intervals may extend into the following year, but the prompt
renders the configured starting year for both endpoints. The displayed end
date can therefore disagree with the date used to compute the label. \\

\rgcode{arithmetic/time_intervals}{time\_intervals}
& For a supported datetime format containing a parenthesized weekday, the
implementation removes everything following the opening parenthesis before
parsing. This also deletes the displayed time and timezone suffix, so the
oracle duration is computed from information different from that shown in
the prompt. \\

\rgcode{logic/aiw}{aiw}
& One default sibling template describes the siblings of a sampled woman but
asks about the brother of an independently sampled man. That man is never
connected to the described family, so the requested sibling count is not
entailed by the instance. \\

\rgcode{geometry/simple_geometry}{simple\_geometry}
& The prompt asserts that the polygon is convex, but generation checks only
that the missing angle is positive. For polygons with at least four sides,
the computed missing interior angle may exceed \(180^\circ\), making the
displayed convex polygon impossible. \\

\rgcode{cognition/rectangle_count}{rectangle\_count}
& The label is the number of rectangle primitives inserted by the generator,
rather than the number of rectangles identifiable in the final ASCII
drawing. Overlapping borders and intersections can form additional complete
rectangles or alternative decompositions not reflected in the stored
count. \\

\addlinespace[4pt]

\addlinespace[4pt]
\multicolumn{2}{l}{\textbf{Material prompt--oracle mismatch}}\\
\addlinespace[2pt]

\rgcode{arithmetic/power_function}{power\_function}
& Negative bases are rendered without parentheses, so an expression such as
\texttt{-2\^{}2} conventionally denotes a value different from
\((-2)^2\). The stored oracle is also emitted as an ordinary Python
floating-point string rather than in the scientific notation explicitly
requested by the prompt. \\

\end{longtable}
\endgroup

\begingroup\footnotesize

\noindent\textbf{Valid but non-canonical targets.}
The following generators retain one sampled witness as the gold answer without
establishing that the rendered instance determines it uniquely:
\rgcode{games/survo}{survo},
\rgcode{code/codeio}{codeio},
\rgcode{cognition/needle_haystack}{needle\_haystack},
\rgcode{algorithmic/sentence_reordering}{sentence\_reordering},
\rgcode{algorithmic/letter_jumble}{letter\_jumble}, and
\rgcode{algorithmic/caesar_cipher}{caesar\_cipher}.
Following the criterion applied to \SL{}, we do not count these as major
semantic defects, since alternative valid witnesses could be admitted by a
suitably general verifier. Unlike the \SL{} cases, however, the shipped
scorers here reject such witnesses outright: \texttt{codeio} never executes
the candidate input, and \texttt{survo} compares cell by cell against the one
sampled matrix. Two further cases are only partly of this kind,
\texttt{letter\_jumble}, also strips punctuation the prompt claims to preserve,
and \texttt{caesar\_cipher} hides the rotation, so the ciphertext is
compatible with every rotation and the target is fixed only by an implicit
English-language prior.
\endgroup

\end{document}